\documentclass{Interspeech}
\usepackage{multirow}
\usepackage{graphicx}
\usepackage{threeparttable}

\interspeechcameraready

\title{Decoding Listener's Identity: Person Identification from EEG Signals Using a Lightweight Spiking Transformer}

\author{Zheyuan Lin$^{1}$, Siqi Cai$^{*,1}$, Haizhou Li$^{1,2,3}$}

\affiliation{School of Data Science}{The Chinese University of Hong Kong}{Shenzhen, China}
\affiliation{Machine Listening Lab}{University of Bremen}{Germany}
\affiliation{Department of ECE}{National University of Singapore}{Singapore}
\email{zheyuanlin@link.cuhk.edu.cn, caisiqi@ieee.org, haizhouli@cuhk.edu.cn}

\usepackage{comment}

\begin{document}

\maketitle

\begin{abstract} 
 EEG-based person identification enables applications in security, personalized brain-computer interfaces (BCIs), and cognitive monitoring. However, existing techniques often rely on deep learning architectures at high computational cost, limiting their scope of applications. In this study, we propose a novel EEG person identification approach using spiking neural networks (SNNs) with a lightweight spiking transformer for efficiency and effectiveness. The proposed SNN model is capable of handling the temporal complexities inherent in EEG signals. On the EEG-Music Emotion Recognition Challenge dataset, the proposed model achieves 100\% classification accuracy with less than 10\% energy consumption of traditional deep neural networks. This study offers a promising direction for energy-efficient and high-performance BCIs. The source code is available at https://github.com/PatrickZLin/Decode-Listener-Identity.
\end{abstract}

\section{Introduction}
Electroencephalography (EEG) serves as a non-invasive method for monitoring and interpreting brain activity, offering insights into cognitive states and enabling the development of brain–computer interfaces (BCIs)~\cite{Yang2024Development}. Within this domain, EEG-based person identification has emerged as a critical challenge, aiming to recognize individuals through their unique neurophysiological signatures~\cite{Yang2022Person}. This capability is crucial for personalized BCI applications, where robust user-specific adaptation is required.

Neuroscience research suggests that auditory stimuli can evoke individualized neural responses~\cite{pandey2022music}. Studies on song identification, stimulus-response relationships, and inter-subject variability during music listening~\cite{warren2008does,vuust2022music} have demonstrated distinct neural patterns influenced by subjective experiences. These findings support the feasibility of identifying listeners based on EEG signals recorded during music engagement.

Traditional machine learning approaches, particularly artificial neural networks (ANNs), have been extensively employed for EEG-based person identification~\cite{Kaur2017Novel,Jayarathne2020Person}. For instance, convolutional neural networks (CNNs) have demonstrated success in user identification tasks due to their ability to extract spatial features from high-dimensional EEG data~\cite{sharma2022neural}. A large-scale study~\cite{Yasaman2024EEGbased} analyzed EEG recordings from 109 subjects using a multilayer fully connected neural network, achieving competitive classification accuracy across all subjects. However, these approaches often rely on handcrafted features (e.g., statistical, frequency, or wavelet descriptors), which limits generalizability. Additionally, ANNs typically require substantial computational resources and may not fully capture the temporal dynamics inherent in EEG signals.

Spiking Neural Networks (SNNs), inspired by the brain's neural architecture, offer a promising alternative due to their event-driven nature and potential for energy-efficient processing. Previous studies have demonstrated the effectiveness of SNNs in EEG classification tasks, which show higher performance with less energy consumption than conventional ANNs~\cite{10220211,faghihi2022neuroscience}. Recent advances in SNNs include models like the Spike-driven Transformer~\cite{yao2024spikedriven}, which integrates transformer mechanisms with spiking neurons to efficiently process sequential data and achieve state-of-the-art performance across various tasks.

This study investigates SNN-based person identification using EEG recordings collected during music listening—a challenging benchmark task that demands biometric identification across different individuals.
To our knowledge, this represents the first exploration of SNNs for EEG-based person identification. We adopt a lightweight spiking transformer, an end-to-end architecture optimized for efficiency and performance. This architecture is designed to efficiently process temporal sequences, making it particularly suitable for handling EEG data, which inherently involves complex temporal dynamics. The results demonstrate SNNs' potential as a robust tool for biometric recognition, advancing the development of more personalized and adaptive BCI systems.

The remainder of this paper is organized as follows: Section 2 outlines the methodology, detailing the spiking transformer architecture for EEG analysis. Section 3 describes the experimental setup, including the EEG dataset specifications, data preprocessing, and training protocol. Results are reported in 
Section 4. Section 5 discusses the implications of our findings and concludes the paper.

\section{Methodology}
\begin{figure}[t]
  \centering
  \includegraphics[width=\linewidth]{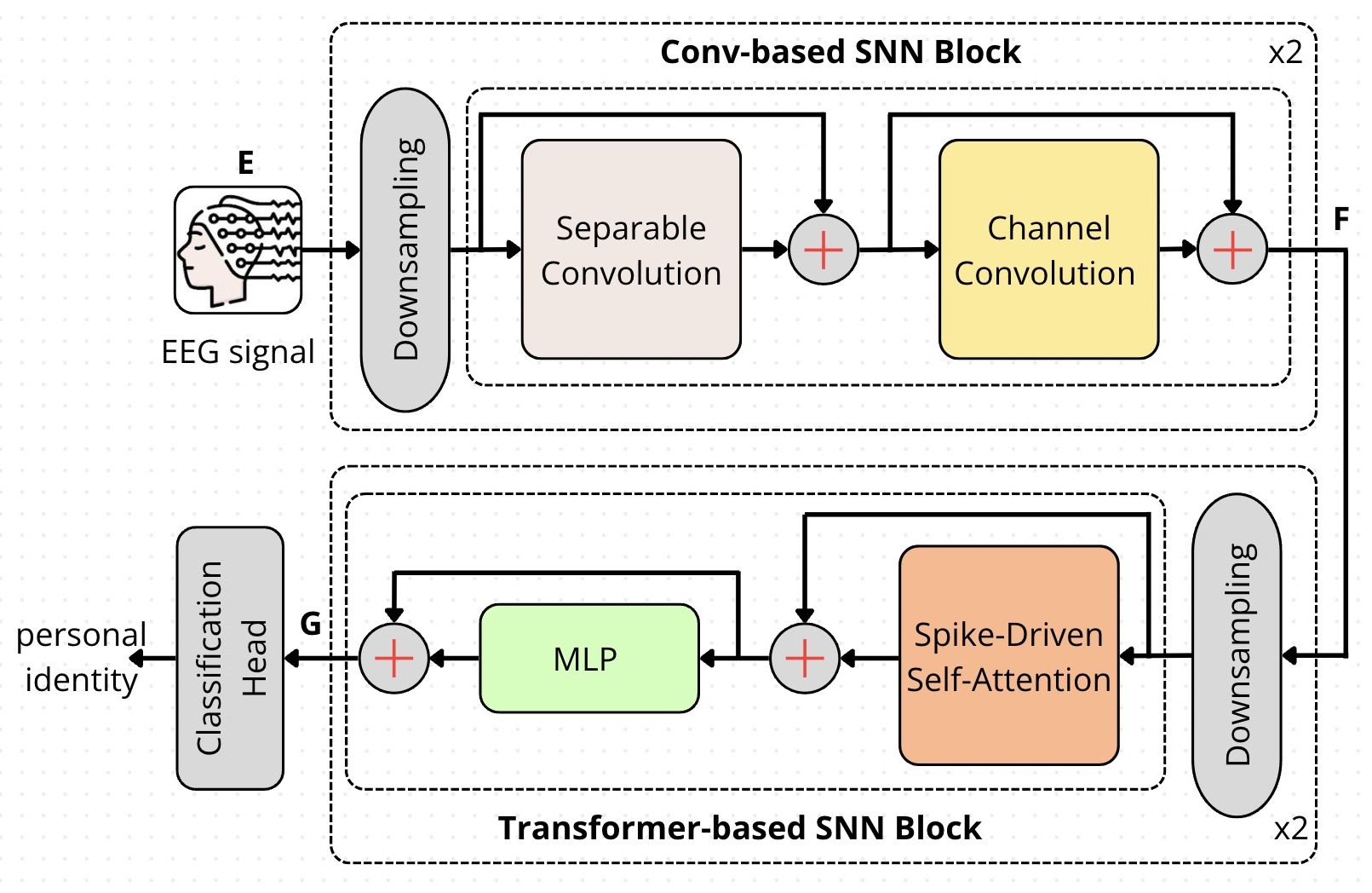}
  \vspace{-0.3cm}
  \caption{Architecture overview of the proposed lightweight spiking transformer for EEG-based person identification. It takes a window of EEG signals as input, and performs multi-class classification to determine the individual's identity. $\bigoplus$ stands for shortcuts.}
  \vspace{-0.2cm}
  \label{fig:model}
\end{figure}

As illustrated in Figure~\ref{fig:model}, the proposed lightweight spiking transformer takes a window of EEG signals as input and performs person identification through multi-class classification.

\subsection{Spiking Neurons}
The core of SNNs is built on spiking neurons, which simulate the behavior of biological neurons more closely than traditional artificial neurons~\cite{tavanaei2019deep}. The spiking neuron layer adopted in our model follows the standard Leaky Integrate-and-Fire (LIF) neuron model~\cite{dayan2001theoretical}.
It integrates temporal and spatial inputs; its membrane potential inherently performs temporal integration, crucial for capturing the dynamics of EEG data. Spikes are generated based on this potential, driving computations forward.
Its synergy can be represented by:
\begin{equation}
V[n+1] = \beta V[n] + \sum_j w_j S_j^{\text{in}}[n] - \text{Reset}[n]
\end{equation}
\begin{equation}
S^{\text{out}}[n+1] = \theta(V[n+1] - \text{V}_{th})
\end{equation}
The membrane potential $V$ of each neuron is updated based on the previous potential, the weighted sum of incoming spikes, and a reset term. $\beta$ represents the leakage factor, $w_j$ are the synaptic weights, and $S_j^{\text{in}}[n]$ denotes the incoming spikes to the neuron at time step $n$. $\text{V}_{th}$ denotes the firing threshold. $\text{Reset}[n]$ ensures that the neuron’s voltage is set to a lower value after firing. A spike is emitted when the membrane potential exceeds a threshold, at which point an output spike $S^{\text{out}}[n+1]$ is activated. This is controlled by $\theta$, the Heaviside step function. 

\subsection{Model Architecture}

\subsubsection{Overall Model}
As illustrated in Figure~\ref{fig:model}, we adopt a lightweight spiking transformer for EEG-based person identification. This architecture is based on the established designs of Spike-driven Transformer V2~\cite{yao2024spikedriven} and Spikformer~\cite{Zhou2022SpikformerWS}. The model combines both Conv-based and Transformer-based SNN blocks to enhance performance and versatility, drawing inspiration from both traditional CNN and Vision Transformer (ViT)~\cite{dosovitskiy2020image} architectures. The first two stages use Conv-based SNN blocks, while the last two leverage Transformer-based SNN blocks.

\subsubsection{Conv-based SNN block}
Separable Convolution (SepConv) is used to extract features from each channel and merge features across channels in the Conv-based SNN blocks. It combines depthwise convolution (kernel size 7x7, stride 1), which performs convolution independently for each input channel, and pointwise convolution, which merges the channels:
\begin{equation}
\text{SepConv}(\textbf{E}') = \text{Conv}_{\text{PW}}\left( \text{Conv}_{\text{DW}}\left( \text{SN}\left( \text{Conv}_{\text{PW}}\left( \text{SN}(\textbf{E}) \right) \right) \right) \right)
\end{equation}
where \textbf{E} represents a window of  EEG signal \( \mathbf{E} \in \mathbb{R}^{C \times T} \), with \( C \) denoting  the number of EEG channels and \( T \) representing the samples within a window. $SN(\cdot)$ represents a layer of LIF neurons as in Section 2.1. $\text{Conv}_{\text{PW}}$ and $\text{Conv}_{\text{DW}}$ denote the pointwise and depthwise convolution operations, respectively.

Channel Convolution combines spatial and channel-wise information, allowing the model to learn deeper, more abstract features. It applies standard convolutions (kernel 3x3, stride 1) to process the output from the SepConv layer, ensuring that spatial patterns are effectively captured:
\begin{equation}
\text{ChannelConv}(\textbf{E}'') = \text{Conv}\left( \text{SN}\left( \text{Conv}\left( \text{SN}(\textbf{E}') \right) \right) \right)
\end{equation}

After the two convolution-based SNN blocks, the input EEG \textbf{E} is transformed into the feature map feature $\textbf{F}$.

\begin{figure}[t]
  \centering
  \includegraphics[width=\linewidth]{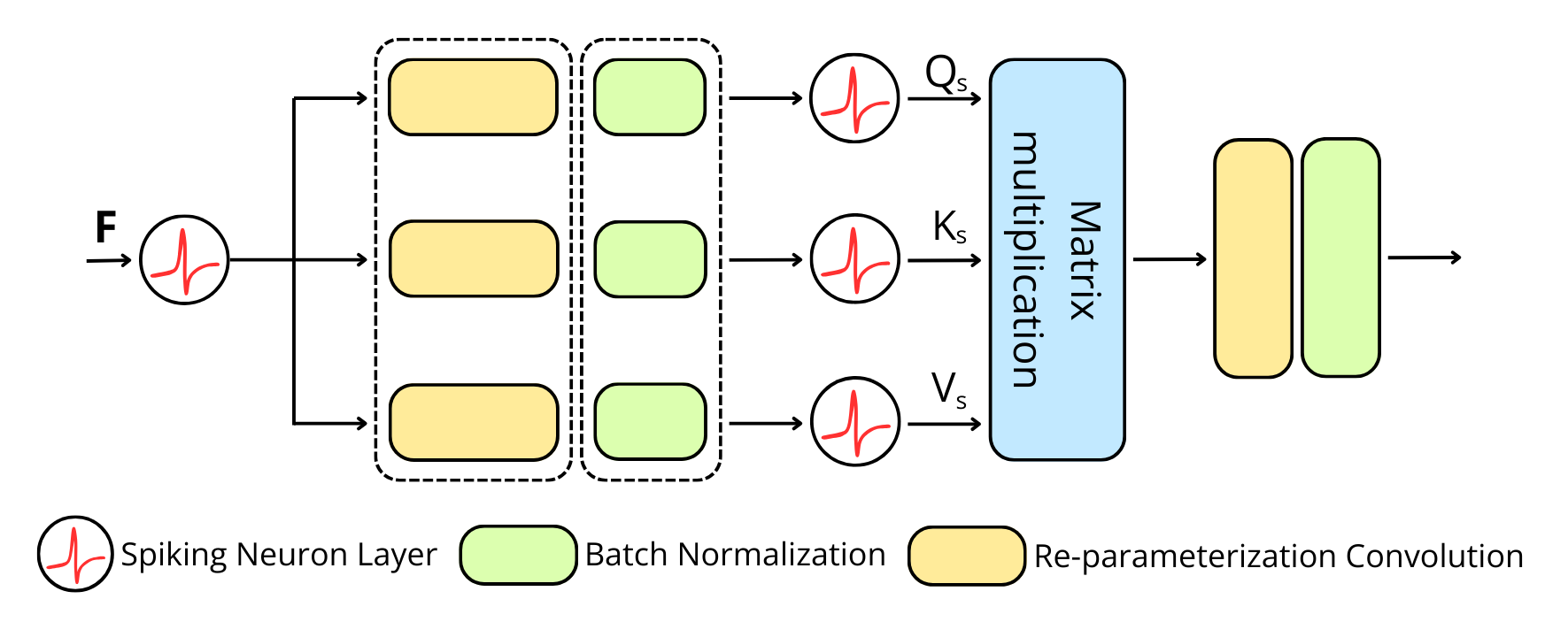}
   \vspace{-0.3cm}
  \caption{Outline of the Spike-Driven Self-Attention (SDSA) block, RepConv stands for re-parameterization convolution.}
  \vspace{-0.2cm}
  \label{fig:SDSA}
\end{figure}

\subsubsection{Transformer-based SNN block}

Spike-driven self-attention (SDSA) replaces the traditional Query (Q), Key (K), and Value (V) matrix operations used in standard attention mechanisms, as illustrated in Figure~\ref{fig:SDSA}. It operates on the spike tensor form Q, K, and V directly, and computes interactions between tokens using sparse additions instead of the dense, computationally expensive dot products found in standard self-attention:
\begin{equation}
\begin{split}
\text{Q}_s &= \text{SN}( \text{RepConv}(\textbf{F}) ) \\ 
\text{K}_s &= \text{SN}( \text{RepConv}(\textbf{F}) ) \\
\text{V}_s &= \text{SN}( \text{RepConv}(\textbf{F}) ) 
\end{split}
\end{equation}
\begin{equation}
\text{SDSA}(\textbf{F}) = \text{RepConv}(\text{SN}(\text{Q}_s (\text{K}_s^T \text{V}_s)))
\end{equation}
where $RepConv(\cdot)$ is re-parameterization convolution (kernel size 3x3, stride 1)~\cite{Ding2021RepVGG}. Unlike conventional Transformers, this avoids expensive operations like softmax and scaling, as spike-form matrix multiplication can be converted to additions using addressing algorithms~\cite{Frenkel2021BottomUpAT}, making it energy-efficient.

Following the SDSA operation, a two-layer MLP is applied to the output. It operates on the channel dimension of the feature maps, improving the model's ability to understand the interactions between different features across channels:
\begin{equation}
\text{MLP}(\textbf{F}') = \text{SN}(\text{SN}(\textbf{F}')W_1)W_2
\end{equation}
where $W_1$ \(\in \mathbb{R}^{C \times rC} \) and $W_2$ \(\in \mathbb{R}^{rC \times C} \) are learnable parameters with MLP expansion ratio r = 4. 

After passing through the two Transformer-based SNN blocks, the feature map $\textbf{F}$ was further modified as \textbf{G}, which serves as the refined feature representation.

Subsequently, $\textbf{G}$ is processed by a linear classification layer, producing the final prediction.

\subsubsection{Parameter Reduction}

To enhance efficiency in EEG-based person identification, we streamlined our model by reducing channel dimensions and optimizing downsampling parameters. Compared to the Spike-driven Transformer V2 baseline~\cite{yao2024spikedriven}, which contains 30.9 million parameters, our approach achieves computational feasibility with just 3.91M parameters while preserving discriminative EEG feature learning and maintaining competitive performance.

Overall, this novel architecture, combining the strengths of spiking neural networks with transformer-based attention mechanisms, enables efficient and accurate processing of EEG signals for person identification tasks. The integration of spiking neurons allows the model to process temporal information in a biologically plausible manner, while the transformer mechanism ensures that long-range dependencies within the data are captured effectively.

\section{Dataset and Experiments}

\subsection{Dataset}
This study utilizes the EEG-Music Emotion Recognition Challenge dataset\footnote{EEG-Music Emotion Recognition Challenge. (n.d.). \textit{EEG-Music Emotion Recognition}. Retrieved February 18, 2025, from https://eeg-music-challenge.github.io/eeg-music-challenge/dataset} for the task of Person Identification—determining the identities of 26 subjects based on EEG responses elicited by musical stimuli.

EEG signals were recorded in a controlled laboratory environment using a 32-channel EPOC Flex system with saline sensors, with data sampled at 128 Hz. Electrode placements followed the international 10-20 system. Each subject participated in 16 trials lasting 90 seconds each. The trials were divided into two categories: personal trials, selected based on the subject's own music preferences, and non-personal trials, chosen randomly from the playlists of other participants.

\subsection{Data Preprocessing}
The EEG signals utilized in this study were obtained from the official EEG-Music Emotion Recognition Challenge dataset, which were underwent preprocessing using an FIR filter (0.5-40 Hz) and artifact removal through Independent Component Analysis (ICA). Moreover, the EEG signals were normalized for each trial within each subject. Each raw EEG trial, approximately 90 seconds in length, was segmented into 10-second windows, each containing 1280 time points.

The training dataset comprises 294 trials collected from 26 subjects, with each subject contributing roughly 12 labeled trials. These trials are used to train models to identify subjects based on their distinct EEG signatures. For validation, a set was extracted from the training data (10\%), ensuring a diverse and representative sample to evaluate model performance during training.

The test dataset consists of 104 held-out trials, all coming from the 26 subjects seen during training. Additionally, 44 extra trials are included in the test set, some of which lack stimulus information or self-assessed emotion labels. This additional data further challenges the model, testing its robustness and generalization capabilities. Person identification results are reported on the testing set.





\subsection{Training Protocol}

The training of the proposed lightweight spiking transformer was conducted over 300 epochs using surrogate gradient\cite{Neftci2019Surrogate}, which facilitates backpropagation through the non-differentiable spiking activity inherent in SNNs. This approach enables efficient learning while preserving the temporal characteristics of the EEG data.

For the loss function, we employed a CrossEntropyLoss with class count balancing to mitigate any potential class imbalances within the dataset. This ensured that the model did not favor any particular subject during training, maintaining fairness and robustness in classification.

Optimization was carried out using the Adam optimizer, chosen for its adaptive learning rate capabilities and overall efficiency in handling sparse gradients. In addition, a ReduceLROnPlateau learning rate scheduler was integrated into the training process. This scheduler dynamically adjusts the learning rate when the validation performance plateaued, thereby enhancing convergence and preventing overfitting.

Together, these components of the training protocol—surrogate gradient-based learning, balanced CrossEntropyLoss, Adam optimization, and adaptive learning rate scheduling—enabled robust and efficient training of the SNN model, culminating in high-performance EEG-based person identification.

\begin{table*}[ht]
\caption{Comparison of different models in terms of person identification performance, parameter count, and energy consumption.}
\label{tab:models}
\centering
\begin{threeparttable}
\begin{tabular}{ccccc}
\hline\hline
Methods  & Architecture  & Parameter count (M) & Energy consumption ($\mu$J) & Accuracy \\ \hline
\multirow{2}{*}{ANN}  
&CCT\cite{Hassani2021EscapingTB}    & 22.0     & 38414.6   & 99.32\%   \\ 
&CCT~\tnote{1}   & 3.24     & 8248.3    & 98.65\%  \\ \hline
\multirow{2}{*}{SNN} &  Spike-driven transformer V2\cite{yao2024spikedriven} & 30.9     & 6065.1         & 100\%    \\ 
 & Ours (Lightweight spiking transformer)      & 3.91     & \textbf{760.7} & \textbf{100\%} \\
\hline\hline
\end{tabular}
\begin{tablenotes}
  \small
  \item[1] CCT model with reduced model size.
\end{tablenotes}
\end{threeparttable}
\vspace{-0.2cm}
\end{table*}

\section{Results}

\subsection{Person Identification Accuracy}
The performance of the models was evaluated using the EEG-Music Emotion Recognition Challenge dataset, with the primary objective of accurately identifying the subject based on EEG signals. In this study, we assessed our proposed lightweight spiking transformer and compared its performance against several other models, including conventional ANNs like the Compact Convolutional Transformer, as well as SNNs such as the Spike-driven Transformer V2~\cite{yao2024spikedriven}. 

\textbf{Compact Convolutional Transformer:} CCT model has demonstrated superior performance in a variety of tasks, outperforming many contemporary transformer models~\cite{Hassani2021EscapingTB}. In this study, we incorporate the CCT model in our experimental setup and further reduce its size to align with the scale of our proposed lightweight transformer.

\textbf{Spike-driven Transformer V2:} Spike-driven Transformer V2 combines transformer-based architecture with spike-driven processing, improving computational efficiency while maintaining high accuracy across various vision tasks~\cite{yao2024spikedriven}. We compared its performance to that of our proposed lightweight spiking transformer to evaluate both efficiency and effectiveness in the task of person identification.

As summarized in Table 1, our lightweight spiking transformer achieved 100\% accuracy in the person identification task. Compared to Spike-driven Transformer V2, which also reached 100\% accuracy, our model significantly reduces the parameter count from 30.9 million to 3.91 million—a nearly 90\% reduction in model size—while maintaining the same level of performance.

In contrast, both variants of the CCT exhibited lower accuracy. The smaller CCT model, despite having a similar parameter count to ours, achieved only 98.65\%. This result further underscores the effectiveness of SNNs in balancing computational efficiency and classification accuracy.

\subsection{Model Efficiency}
We further compare the computational costs of different models, evaluating their inference cost for a single forward pass. The inference cost is measured by the number of floating-point operations per second (FLOPs), based on the standard 45nm CMOS process~\cite{horowitz20141,Yao2023Attention}. Notably, for SNNs, neurons are activated only when the input spikes surpass a threshold, allowing inactive neurons to enter low-power modes, thus are expected to reduce power consumption.

As shown in Table 1, our proposed lightweight spiking transformer, with 3.91 million parameters, demonstrates remarkable energy efficiency, consuming only 760.7 $\mu$J during inference. This low energy consumption can be attributed to the inherent efficiency of our SNN design, which simplifies traditional multiply-accumulate (MAC) operations into more energy-efficient accumulation (AC) operations. Moreover, the SNN operates with an average firing rate of just 7.18\%, meaning that only a small fraction of the network's spiking neurons are active at any given time, further minimizing both computational load and energy usage.

In contrast, the energy consumption of the CCT model was significantly higher than that of the SNN transformer models, with the 22.0 million parameter version consuming 38,414.6 $\mu$J per forward pass. Even the smaller 3.24 million parameter variant, while more efficient, still required 8,248.3 $\mu$J—substantially more than the SNN-based models. Notably, our model consumed only about 10\% of this energy, highlighting its superior efficiency.

In summary, our proposed model distinguishes itself through its exceptional energy efficiency. Its low power consumption, enabled by streamlined computational operations and minimal neuronal firing rates, makes it well-suited for real-time, energy-constrained applications such as BCIs, where both high performance and low energy usage are essential for effective deployment.

The optimized SNN transformer model stands out for its superior energy efficiency. Its low energy consumption, driven by simplified computational operations and low neuronal firing rates, makes it a highly attractive option for real-time, energy-constrained applications such as brain-computer interfaces, where both high performance and low power consumption are crucial.




\subsection{Ablation Study}  
To evaluate the impact of architectural modifications on the spiking transformer, we conducted an ablation study focusing on two critical aspects: (1) downsampling strategy and (2) the role of Conv-based SNN blocks.

\textbf{Downsampling Strategy and Parameter Efficiency:} Compared to Spike-driven Transformer V2, one of the key modifications in our lightweight spiking transformer was the selection of smaller downsampling dimensions. This adjustment reduced the model size from 30.9 million to 3.91 million parameters, significantly lowering computational and energy costs while maintaining 100\% accuracy. The results emphasize the effectiveness of downsampling and parameter pruning in creating a lightweight yet highly accurate model with improved efficiency. Further reductions in downsampling dimensions were explored but led to severe accuracy degradation, indicating that the 3.91 million parameter configuration achieves an optimal balance between efficiency and performance.  

\textbf{Impact of Conv-based SNN Blocks:} To evaluate the necessity of the dual Conv-based SNN blocks in the original architecture, we tested a variant with only one Conv-based block. This further reduced the model size to 2.85 million parameters (a 27\% decrease from the 3.91M version) but resulted in a 2.03\% drop in accuracy. These findings highlight the role of Conv-based SNN blocks in capturing hierarchical temporal features and demonstrate the trade-off between extreme parameter reduction and performance retention.




\section{Discussion and Conclusion}
SNNs offer distinct advantages for EEG-based classification tasks due to their ability to capture the temporal dynamics inherent in neural signals. EEG data is rich with time-dependent patterns, and SNNs, which process information through event-driven, spike-based mechanisms, are particularly well-suited to model these temporal dependencies. This alignment enables SNNs to effectively represent the brain's intricate activity over time, leading to enhanced accuracy in classifying EEG signals.

Another significant benefit of SNNs is their efficiency. The lightweight SNN transformer model in this study achieved high performance while significantly reducing computational and energy costs compared to analogous ANN models. This efficiency is critical for deploying EEG-based systems on edge devices, such as wearables or speech processors, where power consumption and real-time processing are key constraints. The low firing rate of spiking neurons, coupled with simplified operations, makes SNNs particularly suitable for deployment in resource-constrained environments, ensuring that they can operate effectively in portable and energy-efficient settings.

However, despite these promising advantages, several limitations must be considered. While the model performed well on the 26-subject dataset used in this study, its generalizability to larger and more diverse populations remains uncertain. To fully assess the robustness and scalability of the model, further testing with more extensive datasets that incorporate a wider range of demographic and physiological variability is needed. Additionally, future research should investigate the adaptability of SNNs to other EEG-based applications and explore their performance in environments with varying noise levels and signal quality. This would provide a more comprehensive understanding of the potential of SNNs in real-world EEG applications.




\section{Acknowledgements}
This work is supported by National Natural Science Foundation of China (Grant No. 62401377),
internal Project of Shenzhen Research Institute of Big Data, 
internal project of the Guangdong Provincial Key Laboratory of Big Data Computing under the Grant No. B10120210117-KP02, The Chinese University of Hong Kong, Shenzhen (CUHK-Shenzhen),
Shenzhen Stability Science Program 2023, Shenzhen Key Lab of Multi-Modal Cognitive Computing,
Shenzhen Science and Technology Program (Shenzhen Key Laboratory, Grant No. ZDSYS20230626091302006), 
Shenzhen Science and Technology Research Fund (Fundamental Research Key Project, Grant No. JCYJ20220818103001002), 
Program for Guangdong Introducing Innovative and Entrepreneurial Teams, Grant No. 2023ZT10X044,
and Guangdong S\&T Program under Grant 2023B0303040002.

\bibliographystyle{IEEEtran}
\bibliography{main}

\end{document}